\documentclass{article}

\usepackage[nonatbib]{arxiv}
\usepackage[utf8]{inputenc} 
\usepackage[T1]{fontenc}    
\usepackage{hyperref}       
\usepackage{url}            
\usepackage{booktabs}       
\usepackage{amsfonts}       
\usepackage{nicefrac}       
\usepackage{microtype}      
\usepackage{lipsum}
\usepackage{graphicx}

\title{Movement science needs different pose tracking algorithms}

\author{
  Nidhi Seethapathi\thanks{Author for correspondence: snidhi@seas.upenn.edu} \\
  Department of Bioengineering\\
  University of Pennsylvania\\
  Philadelphia, PA 19104 \\
  \texttt{snidhi@seas.upenn.edu} \\
   \And
 Shaofei Wang \\
  Department of Bioengineering\\
  University of Pennsylvania\\
  Philadelphia, PA 19104 \\
  \texttt{sfwang@seas.upenn.edu} \\
 \And
 Rachit Saluja\\
  Department of Electrical Engineering\\
  University of Pennsylvania\\
  Philadelphia, PA 19104 \\
  \texttt{rsaluja@seas.upenn.edu} \\
 \And
 Gunnar Blohm \\
  Center of Neuroscience Studies\\
  Queen's University\\
  Kingston, Ontario \\
  \texttt{gunnar.blohm@queensu.ca} \\
 \And 
 Konrad Paul Kording \\
 Department of Bioengineering\\
  University of Pennsylvania\\
  Philadelphia, PA 19104 \\
  \texttt{kording@seas.upenn.edu } \\
}

\begin{document}
\maketitle

\begin{abstract}
Over the last decade, computer science has made progress towards extracting body pose from single camera photographs or videos. This promises to enable movement science to detect disease, quantify movement performance, and take the science out of the lab into the real world. However, current pose tracking algorithms fall short of the needs of movement science; the types of movement data that matter are poorly estimated. For instance, the metrics currently used for evaluating pose tracking algorithms use noisy hand-labeled ground truth data and do not prioritize precision of relevant variables like three-dimensional position, velocity, acceleration, and forces which are crucial for movement science. Here, we introduce the scientific disciplines that use movement data, the types of data they need, and discuss the changes needed to make pose tracking truly transformative for movement science.

\end{abstract}

\keywords{Movement Science \and Computer Vision \and Pose Estimation}

\section{Movement data matters.}
We only interact with the world through our movements. Consequently, many scientists analyze them. Meaningful analysis of movement data is the key to sports science: good movements maximize performance and minimize the risk of injuries. Movement data is crucial to research in physical and occupational therapy: the right movements improve the quality of life for patients with movement disorders. Quantified movement is a major biomarker for disease: the way people move can aid in diagnosing the disease the patient suffers from. Studying movement is also important in its own right as it is exciting to understand why people move the way they do. Lastly, quantifying movement matters as movement is the output of the brain: movement provides a meaningful goal to be encoded in brain signals. Across all these disciplines, movement data is key.

We begin by highlighting the contributions that movement science research makes to science and medicine across a number of disciplines. Our goal of providing this summary is to introduce computer vision researchers to the importance of developing pose tracking algorithms that  serve movement science well.

\begin{figure*}
    \begin{center}
        \includegraphics{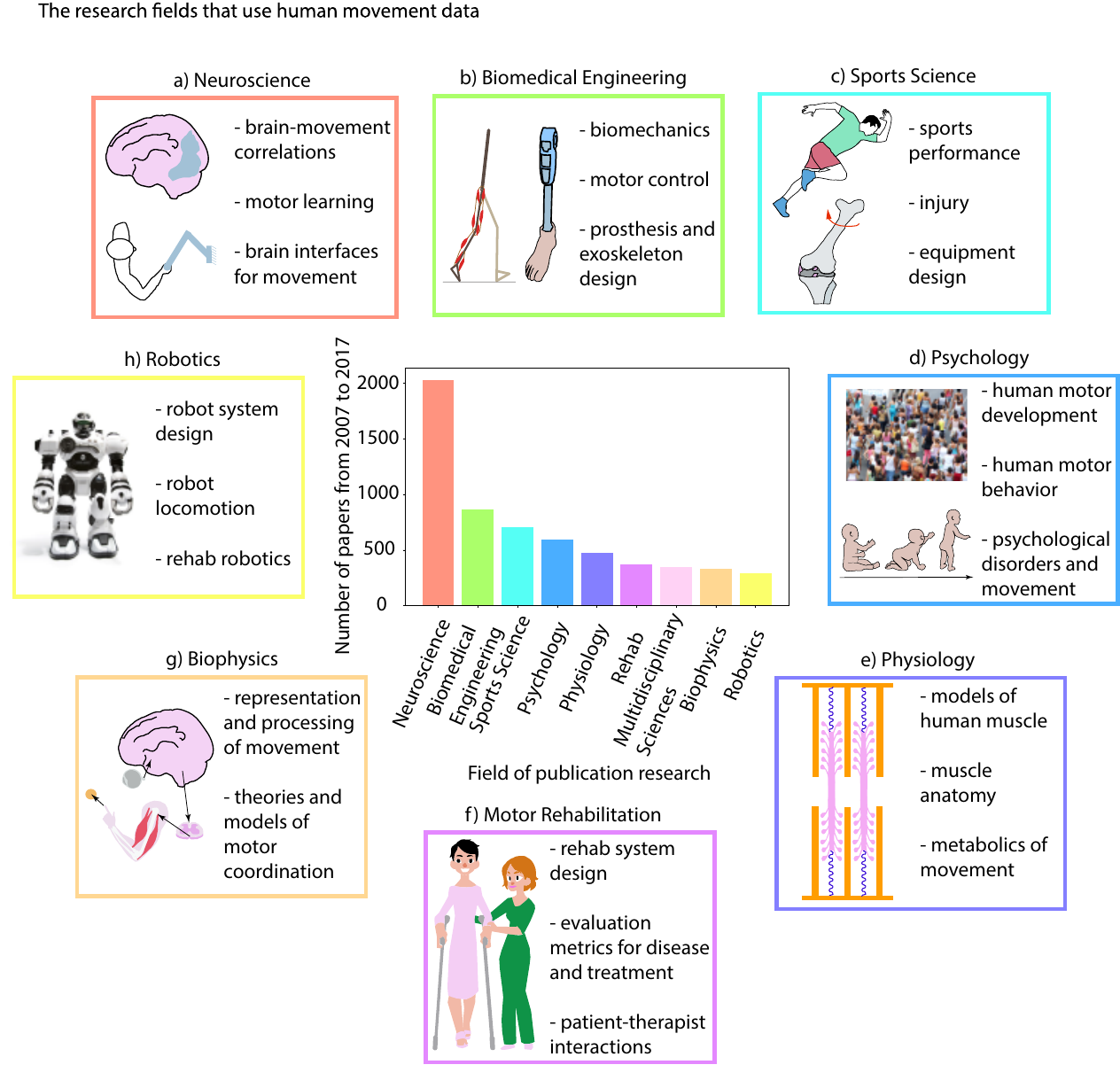}
	        \caption{Many important disciplines of science and engineering rely on human movement data for research.}
	            \label{fig:MovPoseReviewFig1}
	\end{center}
\end{figure*}

\paragraph{Neuroscience.}
As movement is the way an animal interacts with the world, many neuroscientists believe that understanding the neural basis of movement is key to understanding the brain. Thus, many studies investigate how the animal nervous system represents and controls movement (Figure \ref{fig:MovPoseReviewFig1}a). In order to obtain such insights, these studies typically measure the output movement (such as position and velocity) and relate it to measured neural signals. For example, a number of studies investigate how arm reaching movements relate to brain signals. Such studies have found a neural correlate of the preparation to move in a particular direction \cite{churchland2012neural}, the selection of a target to move towards \cite{scherberger2007target} and the inhibition of an impending movement \cite{mirabella2011neural}. These studies lead to an understanding of the neural basis of the generation and control of arm reaching movements. Taken broadly, studying the neural basis of tracked movement helps us understand how the healthy human brain works and points at potential faulty mechanisms in people with brain disorders.

\paragraph{Biomedical engineering.}
In biomedical engineering, movement data is subject to engineering tools, often with the goal of improving health and medicine (Figure \ref{fig:MovPoseReviewFig1}b). These engineers analyze and simulate healthy and diseased human movement (motion and forces) using tools such as multibody dynamics and control theory. For example, many biomedical engineers study human locomotion because it is a ubiquitous daily activity that is crucial for good quality of life. Simulations of locomotion of varying complexity \cite{srinivasan2006computer,delp2007opensim} have been used to analyze diseased walking \cite{skalshoi2015walking}, stability and control of locomotion \cite{seethapathi2019step}, or to understand the effects of lower limb surgery on walking \cite{mansouri2016rectus}. Other engineering research has developed equipment that measure the external forces on the leg \cite{schepers2007ambulatory} and the internal forces on muscles \cite{martin2018gauging} outside the confines of a lab. Thus, biomedical engineering has helped enable the quantification and estimation of movement parameters that are important for human locomotion. More generally, biomedical engineers have used simulation and hardware design to measure and estimate movement data.

\paragraph{Sports and exercise science.}
In sports science, human movement data is often used for maximizing athlete performance and success (Figure \ref{fig:MovPoseReviewFig1}c).  In a typical approach, scientists measure movement features (such as position, velocity, acceleration and force) and correlate them with performance. These results are then used to provide feedback and specific training guidance to athletes in competitive sports. For example, in soccer, the ability to accelerate quickly and accurately has been recognized as important for good performance \cite{spinks2007effects}. Because of this, a number of studies have investigated the effect that different types of training regimes \cite{varley2013acceleration}, ages of players \cite{mendez2011age} and player field positions \cite{taskin2008evaluating} have on acceleration profiles of soccer players' movements. The results from such studies in soccer inform the decisions of coaches when choosing training exercises, selecting players and assigning field positions. More generally, the insights obtained from athlete movement data inform strategy, training, the minimization of injury risk, and eventual success in sports.

\paragraph{Psychology.}
Psychology researchers develop theories of how mental processes (cognition) and perceptions of the environment lead to observable actions; movement is one such action (Figure \ref{fig:MovPoseReviewFig1}d). Such studies measure the kinematics (speed, position, and timing) of movement in response to certain environmental conditions and test theories of mind, senses and body that best explain the observed behavior. For example, a number of studies in psychology analyze which theories best explain locomotion behaviors in different environments. One study found that people walk systematically differently in the presence of music and posited this is due to perceptual amplification \cite{styns2007walking}. Others found that walking behavior in moving crowds have characteristic leader-follower behaviors \cite{rio2014follow} and locomotion in such a moving environment is calibrated in a task-specific way \cite{bruggeman2010direction}. Developmental psychology studies in children also find that locomotion task-environment mapping is learned in a task-specific way \cite{adolph2014fear} and that infants explore their environments through aperiodic short paths and falls \cite{adolph2012you}. Movement serves as a useful quantifiable action that psychologists can analyze to critically evaluate their theories of cognition and perception.

\paragraph{Physiology.}
Physiology-based approaches use movement data to get at the mechanisms of movement. They use measures such as displacements, velocities and forces at the intramuscular, intermuscular or whole-body scales (Figure \ref{fig:MovPoseReviewFig1}e). Additionally, measures of exertion such as muscle activation or metabolic energy consumption are often used. For example, a number of studies in physiology have analyzed how muscle activations are generated and coordinated to control movement and posture. At the intramuscular scale, one study found that motor units within a given muscle are activated over a small spatial range during human standing \cite{vieira2011postural}. Another study analyzed how length change is shared between a tendon and the rest of a muscle in the ankle \cite{loram2007passive} or how muscle lengths change dynamically during walking \cite{cronin2009mechanical}. At the whole-body scale, one study found how multiple muscles are activated in a combined fashion when recovering from a push \cite{torres2007muscle} and another modeled the role of the different muscle sensors in controlling movement \cite{kistemaker2012control}. Through the analysis of movement data across scales, studies in physiology enable mechanistic models of movement.

\paragraph{Rehabilitation.}
About 42 million people in the United States of America are diagnosed with movement disorders such as Parkinson's disease, stroke, and cerebral palsy. Additionally, there are about 2 million amputees in the US and many of them use prostheses everyday. Rehabilitation of people with walking disabilities is a major focus area for all movement science disciplines (Figure \ref{fig:MovPoseReviewFig1}f). Towards this goal, movement data is used as a diagnostic tool, to inform treatment and to quantify the progress of disabled individual post-treatment. For example, movement data is commonly used to diagnose and treat patients with gait disorders. Variability in gait motion is a marker for some movement disorders \cite{schniepp2012locomotion} and can be explained with simple models of fall-avoidance \cite{wang2014stepping}. A proposed dopamine-inducing drug \cite{kurz2010levodopa} and a `deep brain stimulation' treatment \cite{allert2001effects} were found to improve aspects of measured gait in patients with Parkinson's disease. On-line movement measures have been used to design prostheses \cite{wen2019online} and exoskeletons \cite{zhang2017human}, thus paving the way for customizable assistive devices for gait rehabilitation. The quality of medical interventions and diagnostics can be improved by movement quantification.

\paragraph{Biophysics.}
Why humans move the way they do, is one of the big questions in science. Studies investigating this question take inspiration from physics to test theories of biological movement (Figure \ref{fig:MovPoseReviewFig1}g). Typical studies collect movement data to test hypotheses such as minimization of metabolic energy \cite{alexander1997minimum}, optimal feedback control \cite{liu2007evidence}, or Bayesian inference \cite{kording2007causal}. For example, scientists ask if the way people walk is as efficient as it could be. They do so by measuring metabolic energy use and kinematics and relating the two through optimization models \cite{srinivasan2006computer,ackermann2010optimality}. Minimization of metabolic energy predicts observed walking speed \cite{seethapathi2015metabolic}, step width \cite{maxwell2001mechanical} and step frequency \cite{bertram2001multiple}. Scientists have also studied the relationship between metabolic energy and stability i.e., walking while minimizing the chances of a fall \cite{bruijn2009slow,dean2007effect}. Movement data thus enables answering questions about the how and why of human movement.

\paragraph{Robotics.}
Inspiration from human movement has been used to develop more stable and more efficient robot motion (Figure \ref{fig:MovPoseReviewFig1}h). Moreover, there is a recent thrust in the field of robotics to assist with physical therapy in a more repeatable, quantifiable and low-cost environment. Towards this goal, robots use movement data as a source of inspiration for mechanical design, to inform robot control algorithms and to act as a goal signal (in the case of rehab robots). For example, one robot project takes inspiration for leg mechanical design and control from birds with an inverted knee \cite{vejdani2013bio}. While most robots consume a lot of power for locomotion \cite{chestnutt2007locomotion}, some robots take inspiration from the mechanics of human locomotion to achieve energy-efficient motion \cite{bhounsule2012design}. Robots intended for rehabilitating people with walking disorders such as wearable soft exoskeletons \cite{ding2014multi} and low-cost interactive robots that train and monitor patient progress \cite{johnson2007potential} are at the cutting edge of the field. Thus, human and animal movement serves as a source of conceptual inspiration as well as provides data for designing and controlling useful robot movement.

\paragraph{Other disciplines that use movement data.}
The above description of the disciplines that use movement data is by no means an exhaustive one. Our focus in this review is biased by the expertise of the authors towards scientific and medical applications that aim to understand, assist, and improve human movement. Social sciences, comparative biology, security applications, etc. are examples of fields that also use human and animal movement data and are outside the scope of our expertise. For example, in social sciences, human movement data is used as a metric of body language to infer emotion \cite{barliya2013expression} with applications to animation \cite{hicheur2013perception} and social robotics \cite{lourens2010communicating}. Measuring  movements of different animal species as a function of morphology is of interest to comparative biologists interested in understanding the relationship between morphology and movement behavior \cite{more2010scaling,usherwood2008compass}. In the field of security, analyzing human movement is important to detect stealthy or threatening behavior using security cameras \cite{lin2011human,neverova2016learning}. Thus, our vision for how pose tracking needs to change in order for it to transform movement science may extend to disciplines beyond the specific ones mentioned in this paper.

\section{Pose tracking promises to transform the scope and scale of movement science.}

Movement science is an interdisciplinary field that has impacted medicine, engineering, neuroscience and sports with thousands of papers being published (see Figure \ref{fig:MovPoseReviewFig1}) and many tens of thousands being cited every year. However, the traditional tools used for data collection in movement science significantly limit the scope and scale of its study. A vast majority of movement science focuses on contrived and repetitive movements studied inside the confines of a lab, is conducted on small non-representative subject samples and uses expensive equipment for measurements. Computer vision-based tracking of human pose, if it meets its potential, promises to transform movement science \cite{wei2018behavioral} by broadening its scope, increasing its scale, making it more representative and less expensive.

 A majority of movement science studies are conducted inside the confines of a lab, as demanded by large and tethered sensors. This prevents the community from studying the broad range of human movement behavior found in the real world. Also, as lab-confined studies can be quite time consuming, they are often conducted on small sample sizes of human subjects. High quality computer vision-based pose tracking of videos promises to capture complex human movements occurring in natural environments \cite{chambers2019pose}. Also, because it is relatively easy to obtain videos of human movement from online sources (such as YouTube) with the help of creative commons license, pose tracking could increase the sample sizes of data used in movement science studies by orders of magnitude. 

The high cost of the sensors used to measure movement data and the difficulty of recruiting subjects widely limits the accessibility and inclusivity of the science. A substantial fraction of movement science studies are conducted largely on American males of a standard height, size and age as these subjects are easily available in a university setting; this fact is likely to bias the findings of movement science towards a subset of the population. Pose tracking provides the opportunity to conduct studies on people of all sexes, shapes, sizes, and ages; this promises to make the research findings more broadly-applicable and generalizable. Most existing tools for movement science are very expensive and limit the science only to labs that have a lot of funds available. Pose tracking research has the ability to bring movement science to labs that have less access to resources such as smaller schools and departments in developing countries.

There is constant ongoing research to improve the state of the art of computer vision-based pose tracking tools. Despite this, the current pose tracking algorithms fall short of their potential to transform movement science. In the next section, we outline some of the reasons why current pose tracking algorithms are unsuitable to the needs of movement science and suggest some of the ways in which they can be changed to transform movement science.

\section{Pose tracking algorithms need to estimate different movement quantities.}

Movement science needs good estimates of three-dimensional kinematics, mass, size and kinetics of human and animal movement. For example, body part positions matter for muscle and tendon lengths. Velocities matter for neural signals. Accelerations matter for animals chasing one another. Forces matter for injuries. Energy matters for efficiency of movement. While the two-dimensional position of the left ear of an athlete on a single image may be perfectly scientifically irrelevant, the force of impact on her leg may decide between an outstanding career in baseball and an outstanding bill for physical therapy. Despite this, two-dimensional pixel positions are popular in computer vision as they are easy to obtain and many competitions have been dedicated to maximizing their estimation accuracy. In the rest of this section, we argue that pose tracking should start working to improve the estimation of the quantities that actually matter for doing science with movement data.

\begin{figure*}
    \begin{center}
        \includegraphics{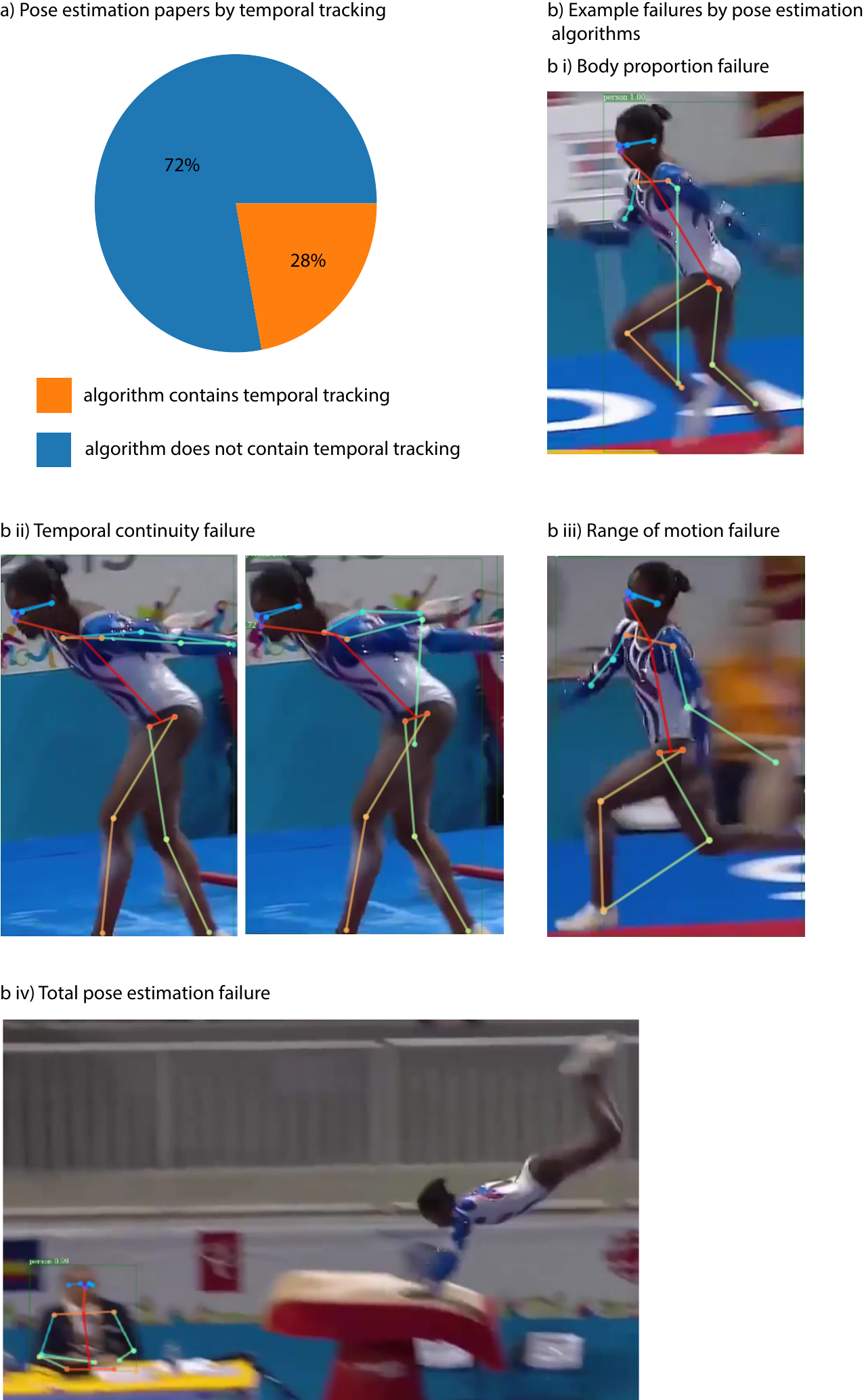}
	        \caption{\textbf{The need for better pose tracking algorithms.} a) Most pose estimation papers published in computer vision conferences in the past two years do not use temporal information. b) Typical failure modes when algorithms are applied to videos of interest to movement science. In these cases, the algorithm's performance is clearly inferior to that of the human eye. To generate this figure, we processed a video of a gymnast using the pretrained keypoint-RCNN model from Detectron \cite{Detectron2018}. Example images (CC) taken from YouTube, https://youtu.be/C08QnMjDfIc?t=79}
	            \label{fig:MovPoseReviewFig2}
	\end{center}
\end{figure*}

Current pose tracking algorithms do not prioritize measurement of the quantities that matter for movement science. The major focus of pose estimation research so far has been on estimating 2D pose from single images; the focus of the field is now quickly moving towards 3D pose from single images. However, in most pose estimation algorithms (see Figure \ref{fig:MovPoseReviewFig2}a), consecutive time frames are treated as statistically independent and the underlying dynamical structure of the pose statistics are ignored. This omission often results in gross mis-estimates of pose in consecutive frames of a video (for instance, see Figure \ref{fig:MovPoseReviewFig2}b ii) that can easily be discerned by the human eye. In addition to not incorporating structure in time into the algorithms, the ground truth data benchmarks currently used do not include quantities important to movement science like velocity, acceleration, and forces \cite{ionescu2013human3,lin2014microsoft}; such benchmarks often measure keypoint localization errors averaged over all frames. Moreover, these benchmark datasets do not consist of the types of movements encountered in movement science, often consisting of contrived poses \cite{ionescu2013human3} or relying on much broader-purpose image datasets \cite{ionescu2013human3}. In a popular variant of pose tracking, multi-person pose tracking \cite{andriluka2018posetrack}, the community has focused more on identity-switches and fragmentations of multiple targets, giving even less attention for localization accuracy, not to mention metrics such as velocity and acceleration. Thus, we believe that the field of pose estimation currently does not prioritize important movement variables and this results in poor estimates of the data that matters for movement science. 

\begin{figure*}
\begin{center}
        \includegraphics[scale=0.35]{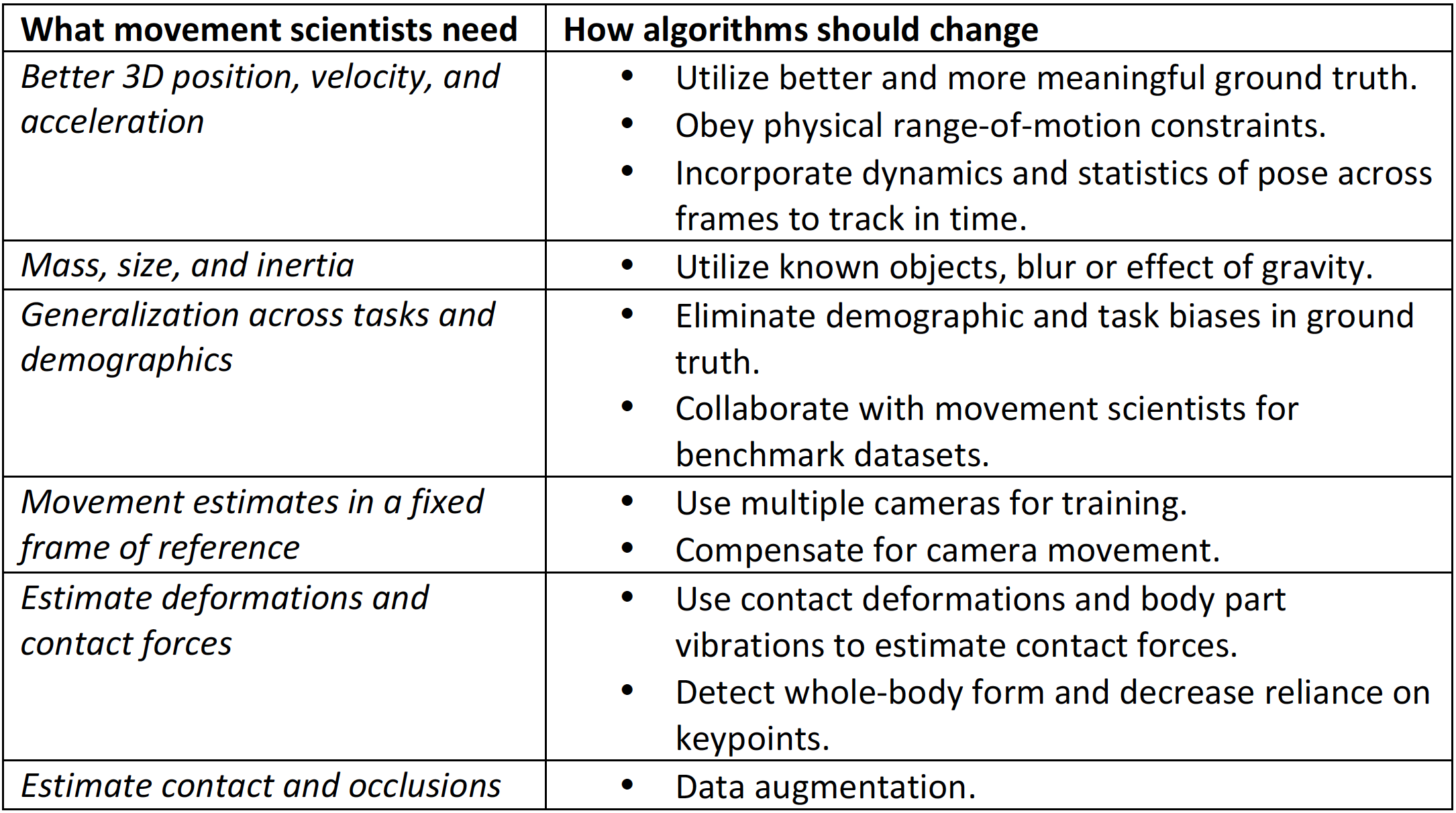}
	        \caption{\textbf{Key takeaways from the paper regarding what movement science needs from pose tracking and how to get there.}}
	            \label{fig:MovPoseReviewFig3}
	   \end{center}
\end{figure*}

Some common failure modes of existing pose estimation algorithms are illustrated in Figure \ref{fig:MovPoseReviewFig2}b. In this section, in addition to highlighting the quantities that matter to movement science, we suggest ways that pose tracking algorithms should be adapted to better estimate these quantities. We provide a tabular summary of our key suggestions to the computer vision community in Figure \ref{fig:MovPoseReviewFig3}.

\paragraph{Three-dimensional position, velocity and acceleration.}
Our movements unfold in three dimensions and most movement science studies focus on three-dimensional positions, velocities, and accelerations. For example, to diagnose progress in a patient with a movement disorder, measures of the three-dimensional kinematics are analyzed \cite{hong2009kinematic}. A majority of the existing pose tracking algorithms, however, aim to maximize the accuracy of two-dimensional, not three-dimensional, pose in single images or video frames \cite{sun2019deep,yang2017learning,newell2016stacked,yang2018parsing}. While more and more pose tracking algorithms are recently aiming for three-dimensional estimates \cite{kocabas2019self,sun2018integral}, still fewer incorporate tracking in time to improve pose estimates i.e. using the past and future movements to improve localization of pose in a given frame (Figure \ref{fig:MovPoseReviewFig2}a). Frame-to-frame tracking errors of the kind shown in Figure \ref{fig:MovPoseReviewFig2}b ii will lead to even larger errors in velocity and acceleration upon numerical differentiation.

Despite the existing issues, we believe that obtaining more accurate three-dimensional positions, velocities and accelerations from videos is possible by changing the pose tracking algorithms and ground truth benchmarks. Skeletal motion naturally creates a hierarchical dependency structure that results in spatial (joint location) and temporal (laws of motion) constraints. Thus, if you know the position, velocity, acceleration, and skeletal shape from the past few frames, then you can build a strong prior for the next frame. None of the existing pose tracking algorithms use such priors to improve pose estimates. Secondly, it is possible to incorporate into the algorithm camera motion to obtain 3D depth information \cite{zhou2017unsupervised,vijayanarasimhan2017sfm}. In addition to the issues with existing algorithms outlined above, pose estimation algorithms use crowd-sourced hand-labeled keypoints \cite{alp2018densepose,lin2014microsoft} as ground truth and these are likely subject to human error \cite{goodman2013data}. Also, 3D pose tracking algorithms use contrived lab-based poses for ground truth \cite{ionescu2014human3} that likely do not overlap with the distribution of poses that are of interest to movement scientists. Video ground truth data (not static images) for velocity and acceleration in-the-wild could be obtained by collaborating with movement scientists. Then, minimization of errors in velocity and acceleration, not just pose, can be added to the objective functions of the algorithms. Given that many approaches in movement science are infeasible without reasonably accurate three-dimensional kinematic measures beyond static pose, pose tracking algorithms must improve their estimation of three-dimensional movements.

\paragraph{External contact forces.}
For many applications, quantifying the external forces involved in a movement is important. After all, the external forces determine the stress on bones and joints which relate to injury \cite{schache2009biomechanical}. However, estimating external forces with current pose tracking algorithms is practically impossible. If there is only one point of contact, one can estimate contact forces using the mass and acceleration estimates for the individual body segments. However, when there are multiple points of contact for a movement, forces cannot be directly estimated from mass and acceleration because such a system is `statically indeterminate' \cite{chao1978graphical}. Moreover, the relevant frequency content of even relatively slow movements such as walking goes up to about 20 Hz \cite{stergiou2002frequency} which, according to Nyquist's theorem \cite{Weik2001}, cannot be observed with the frame rate of a typical video camera (about 30 Hz).

This drawback of existing algorithms, however, does not mean that a video-based algorithm cannot successfully estimate external force from movement. Consider someone kicking a ball: their foot decelerates by some extent (which can be estimated from previous and subsequent frames) over the total displacement of the soft tissue of the foot, resulting in the deformation of the shoe-ball complex and the movement of the ball \cite{shinkai2008ball}. In addition to local deformation near the contact patch, estimating transmitted vibrations of individual body parts (the calf of the leg, for instance) can also help break static indeterminacy and estimate contact forces. As the deformations over which forces occur and the vibrations they result in contain information about the forces themselves, an algorithm designed to estimate external forces from videos should be achievable. Given the importance of external force measurements for movement science, their estimation should be prioritized by the pose tracking algorithms and competitions.

\paragraph{Absolute mass, length and inertia.}
Estimates of true whole-body as well as body segment mass and size are necessary for movement science. Estimation of mass and size of individuals is important for understanding how movements differ in people of different body types \cite{mcmanus2010children} and most movement science papers are expected to report the weight and height statistics of the subjects studied. The mass and inertia measures of body segments are needed to estimate internal forces and torques and to identify the joints that contribute to a given movement \cite{ren2008whole}. Despite this, typical pose tracking algorithms do not attempt to provide estimates of absolute mass, inertia and size. The best one could do with existing pose tracking algorithms is to obtain the relative size of one segment with respect to other ones. However, such relative measures of size are typically not useful: absolute measures of movement are needed for any type of diagnosis and when comparing movements across individuals of different sizes.

One way to rectify this is by using computer vision to estimate the true size and scale of a known object in the background and use this information to estimate the size of the person or animal in the image. The estimates of absolute mass and inertia can then be made using standard cadaveric length-mass and length-inertia regressions \cite{dumas2007adjustments}. Adding in priors for the relative sizes of different body segments from empirical data will also prevent errors in pose estimation which result in impossible body lengths (see Figure \ref{fig:MovPoseReviewFig2}b i). One could also use optical effects, such as limited depth of field and their influence on blur to estimate the true size of the object in a given image. Gravity is a constant and affects the dynamics of freely falling objects seen in videos, this information can be used to estimate the true size of a given object based on the direction of gravity and the trajectories inferred frame-by-frame. This could be done, for instance, by tuning the scaling between the distance in pixels and the true length until the vertical acceleration of the object is equal to acceleration due to gravity. One might also need to incorporate or ignore the effect of air resistance depending on the application. Absolute size and scale of movement is essential for making meaningful scientific inference from movement data and pose tracking algorithms should estimate these.

\paragraph{Pose tracking with task and subject generality.}
Aspects of movement have been found to change with subject demographics (like age or sex) and with the movement task (like walking or reaching). For example, studying the likelihood of falls with age \cite{hollman2007age} is an important area of movement science as many disabilities and deaths in older individuals occur due to falls \cite{akyol2007falls}. Also, the differences in the injury-proneness of movements in males and females has been studied \cite{liederbach2014comparison}. However, it is unclear if pose tracking algorithms will work equally well on people of different demographics or across different movement tasks. Artificial intelligence has been shown to have racial and gender biases due to unbalances datasets \cite{osoba2017intelligence}. Providing labels for demographic information and for the type of movement task could help study any systematic biases in the pose estimates. For example, infants and elderly individuals have different body configurations than the rest of the population. Infants have a comparatively larger torso and head while elderly individuals often have a hunched posture and use walkers or crutches; these body configurations are less typically seen in the image datasets used to train pose tracking models. In Figure \ref{fig:MovPoseReviewFig2}b iv, we show an example of this where the pose estimation algorithms completely fails to detect an upside-down gymnast while still detecting a sitting human with similar amounts of blur, likely due to not being exposed to training data that contains a gymnast mid-task. Thus, it is very likely that the current pose tracking algorithms would need significant retraining to be able to correctly detect the movements of the populations that are of interest for movement science.

Pose tracking algorithms could be better designed to detect the movements of a broader demographic by training the models on a more diverse range of videos that include elderly individuals \cite{robinovitch2013video}, infants \cite{karayiannis2001extraction}, etc. Also, they could be trained to generalize across different tasks by training with videos that contain different types of movements that are of interest to movement science like walking, running, reaching, etc. Computer vision can also provide an estimate of demographics, say, by classifying the face by age \cite{yi2014age} or sex \cite{xu2008hybrid}. Additionally, these algorithms could also be trained to classify demographics based on the movements themselves. Inferring the demographics of the individuals and the type of movement task in addition to tracking pose is essential to study movements in different populations, to ensure unbiased training datasets, and to remove any systematic biases in the estimates for certain populations and tasks.

\paragraph{Body contact and partial occlusion.}
Pose tracking promises movement science on large datasets in-the-wild which will be useful for many applications that are otherwise difficult to study. However, many in-the-wild environments that are of interest to movement science consist of body contact between multiple individuals and partial occlusion of body parts. For example, studying physiotherapist-patient interactions in a clinical setting necessitates separating the movements of the physiotherapist from those of the patient despite contact \cite{mendonca2018quantifying}, something current pose tracking algorithms would not be good for. Similarly, in-the-wild data consists of occlusions which currently cause some pose tracking algorithms to fail \cite{cao2018openpose} with the only solution being to handpick video frames where occlusions are absent. 

These issues with not detecting contact and partial occlusions can be dealt with, for instance, by creating training examples by augmenting ground truth data such that individuals from distinct images are artificially brought in contact or occluded, to train the algorithms to better detect such scenarios. The use of such synthetic data and data augmentation will also provide more accurate ground truth data than the currently used hand-labeled estimates of the location of an occluded body part. Additionally, this approach will help balance the dataset by creating more examples that contain occlusion and contact in comparison to the datasets that are currently used. By making pose tracking algorithms better at detecting partial occlusions and contact, we can truly leverage the wealth of in-the-wild data to answer movement science questions.

\paragraph{Fixed frame of reference.}
For many movement science applications, the estimates of body motion need to be in a fixed frame of reference. For example, what matters for understanding the progress of an individual undergoing physiotherapy is the knee angle in the body's frame of reference \cite{joukov2014online} not in the camera frame of reference. Moreover, Newton's laws of motion, which are always used for movement science analyses, only hold true in a frame of reference that is static or moving at a constant speed. Current pose tracking methods would estimate such body angles in two-dimensions in the camera's frame of reference. However, this method is subject to hand-held camera movements \cite{delbracio2015burst} and the angles estimated would depend heavily on the camera angle, which is not ideal. 

One way to deal with the issue of camera-fixed frame of reference in hand-held videos is to use training data where multiple camera angles for the same movement are naturally present, e.g. during sporting events.  Additionally, pose tracking algorithms could be trained to use a fixed background object to estimate and remove camera movements from the pose estimates, say, by using SLAM and self calibration to update camera pose \cite{mur2015orb}. Being able to provide good movement estimates in a fixed frame of reference is crucial for movement science applications.

\section{Conclusions}
The pose tracking field has made dramatic progress over the course of the last decade. And, indeed, there are impressive demonstrations that show how great the technology is \cite{cao2018openpose,alp2018densepose}. However, the field has not yet, with the exception of DeepLabCut \cite{mathis2018deeplabcut,nath2019using}, impacted movement science research because its algorithms do not prioritize the quantities that matter for movement science. In this paper, we have introduced computer vision scientists to the field of movement science, outlined the reasons why computer vision has failed to impact movement science despite the obvious scope for connections, and outlined some of the ways in which pose estimation algorithms can be adapted to bridge this gap. We believe that it is time to design pose tracking algorithms around the needs of the community that actually needs pose tracking: movement science.

\paragraph{Acknowledgments.}
We thank Claire Chambers for her comments on the issues she faced when using existing pose estimation software.
\paragraph{Funding Statement.}
This work was funded by NIH grant R01NS063399.

\paragraph{Competing Interests.} The authors declare that they have no competing interests.

\paragraph{Authors' Contributions.}
NS conceived, wrote and edited the paper, and created the figures and table. SW helped generate figure 2. RS and GB provided comments on the paper. KPK conceived the purpose and scope of the paper and provided ideas. All authors edited the paper.

\bibliographystyle{unsrt}
\bibliography{references}

\begin{thebibliography}{10}

\bibitem{churchland2012neural}
Mark~M Churchland, John~P Cunningham, Matthew~T Kaufman, Justin~D Foster, Paul
  Nuyujukian, Stephen~I Ryu, and Krishna~V Shenoy.
\newblock Neural population dynamics during reaching.
\newblock {\em Nature}, 487(7405):51, 2012.

\bibitem{scherberger2007target}
Hansj{\"o}rg Scherberger and Richard~A Andersen.
\newblock Target selection signals for arm reaching in the posterior parietal
  cortex.
\newblock {\em Journal of Neuroscience}, 27(8):2001--2012, 2007.

\bibitem{mirabella2011neural}
Giovanni Mirabella, Pierpaolo Pani, and Stefano Ferraina.
\newblock Neural correlates of cognitive control of reaching movements in the
  dorsal premotor cortex of rhesus monkeys.
\newblock {\em American Journal of Physiology-Heart and Circulatory
  Physiology}, 2011.

\bibitem{srinivasan2006computer}
Manoj Srinivasan and Andy Ruina.
\newblock Computer optimization of a minimal biped model discovers walking and
  running.
\newblock {\em Nature}, 439(7072):72, 2006.

\bibitem{delp2007opensim}
Scott~L Delp, Frank~C Anderson, Allison~S Arnold, Peter Loan, Ayman Habib,
  Chand~T John, Eran Guendelman, and Darryl~G Thelen.
\newblock Opensim: open-source software to create and analyze dynamic
  simulations of movement.
\newblock {\em IEEE transactions on biomedical engineering}, 54(11):1940--1950,
  2007.

\bibitem{skalshoi2015walking}
Ole Skalsh{\o}i, Christian~Hauskov Iversen, Dennis~Brandborg Nielsen, Julie
  Jacobsen, Inger Mechlenburg, Kjeld S{\o}balle, and Henrik S{\o}rensen.
\newblock Walking patterns and hip contact forces in patients with hip
  dysplasia.
\newblock {\em Gait \& posture}, 42(4):529--533, 2015.

\bibitem{seethapathi2019step}
Nidhi Seethapathi and Manoj Srinivasan.
\newblock Step-to-step variations in human running reveal how humans run
  without falling.
\newblock {\em eLife}, 8:e38371, 2019.

\bibitem{mansouri2016rectus}
Misagh Mansouri, Ashley~E Clark, Ajay Seth, and Jeffrey~A Reinbolt.
\newblock Rectus femoris transfer surgery affects balance recovery in children
  with cerebral palsy: a computer simulation study.
\newblock {\em Gait \& posture}, 43:24--30, 2016.

\bibitem{schepers2007ambulatory}
H~Martin Schepers, Hubertus~FJM Koopman, and Peter~H Veltink.
\newblock Ambulatory assessment of ankle and foot dynamics.
\newblock {\em IEEE Transactions on Biomedical Engineering}, 54(5):895--902,
  2007.

\bibitem{martin2018gauging}
Jack~A Martin, Scott~CE Brandon, Emily~M Keuler, James~R Hermus, Alexander~C
  Ehlers, Daniel~J Segalman, Matthew~S Allen, and Darryl~G Thelen.
\newblock Gauging force by tapping tendons.
\newblock {\em Nature communications}, 9(1):1592, 2018.

\bibitem{spinks2007effects}
Christopher~D Spinks, Aron~J Murphy, Warwick~L Spinks, and Robert~G Lockie.
\newblock The effects of resisted sprint training on acceleration performance
  and kinematics in soccer, rugby union, and australian football players.
\newblock {\em The Journal of Strength \& Conditioning Research}, 21(1):77--85,
  2007.

\bibitem{varley2013acceleration}
Matthew~C Varley and Robert~J Aughey.
\newblock Acceleration profiles in elite australian soccer.
\newblock {\em International journal of sports medicine}, 34(01):34--39, 2013.

\bibitem{mendez2011age}
Alberto Mendez-Villanueva, Martin Buchheit, Sami Kuitunen, Andrew Douglas, Esa
  Peltola, and Pitre Bourdon.
\newblock Age-related differences in acceleration, maximum running speed, and
  repeated-sprint performance in young soccer players.
\newblock {\em Journal of sports sciences}, 29(5):477--484, 2011.

\bibitem{taskin2008evaluating}
Halil Taskin.
\newblock Evaluating sprinting ability, density of acceleration, and speed
  dribbling ability of professional soccer players with respect to their
  positions.
\newblock {\em The Journal of Strength \& Conditioning Research},
  22(5):1481--1486, 2008.

\bibitem{styns2007walking}
Frederik Styns, Leon van Noorden, Dirk Moelants, and Marc Leman.
\newblock Walking on music.
\newblock {\em Human movement science}, 26(5):769--785, 2007.

\bibitem{rio2014follow}
Kevin~W Rio, Christopher~K Rhea, and William~H Warren.
\newblock Follow the leader: Visual control of speed in pedestrian following.
\newblock {\em Journal of vision}, 14(2):4--4, 2014.

\bibitem{bruggeman2010direction}
Hugo Bruggeman and William~H Warren.
\newblock The direction of walking—but not throwing or kicking—is adapted
  by optic flow.
\newblock {\em Psychological Science}, 21(7):1006--1013, 2010.

\bibitem{adolph2014fear}
Karen~E Adolph, Kari~S Kretch, and Vanessa LoBue.
\newblock Fear of heights in infants?
\newblock {\em Current directions in psychological science}, 23(1):60--66,
  2014.

\bibitem{adolph2012you}
Karen~E Adolph, Whitney~G Cole, Meghana Komati, Jessie~S Garciaguirre, Daryaneh
  Badaly, Jesse~M Lingeman, Gladys~LY Chan, and Rachel~B Sotsky.
\newblock How do you learn to walk? thousands of steps and dozens of falls per
  day.
\newblock {\em Psychological science}, 23(11):1387--1394, 2012.

\bibitem{vieira2011postural}
Taian~MM Vieira, Ian~D Loram, Silvia Muceli, Roberto Merletti, and Dario
  Farina.
\newblock Postural activation of the human medial gastrocnemius muscle: are the
  muscle units spatially localised?
\newblock {\em The Journal of physiology}, 589(2):431--443, 2011.

\bibitem{loram2007passive}
Ian~D Loram, Constantinos~N Maganaris, and Martin Lakie.
\newblock The passive, human calf muscles in relation to standing: the
  non-linear decrease from short range to long range stiffness.
\newblock {\em The Journal of physiology}, 584(2):661--675, 2007.

\bibitem{cronin2009mechanical}
Neil~J Cronin, Masaki Ishikawa, Michael~J Grey, Richard~Af Klint, Paavo~V Komi,
  Janne Avela, Thomas Sinkjaer, and Michael Voigt.
\newblock Mechanical and neural stretch responses of the human soleus muscle at
  different walking speeds.
\newblock {\em The Journal of physiology}, 587(13):3375--3382, 2009.

\bibitem{torres2007muscle}
Gelsy Torres-Oviedo and Lena~H Ting.
\newblock Muscle synergies characterizing human postural responses.
\newblock {\em Journal of neurophysiology}, 2007.

\bibitem{kistemaker2012control}
Dinant~Arne Kistemaker, Arthur Knoek~J Van~Soest, Jeremy~D Wong, Isaac~L
  Kurtzer, and Paul~L Gribble.
\newblock Control of position and movement is simplified by combined muscle
  spindle and golgi tendon organ feedback.
\newblock {\em American Journal of Physiology-Heart and Circulatory
  Physiology}, 2012.

\bibitem{schniepp2012locomotion}
Roman Schniepp, Maximilian Wuehr, Maximilian Neuhaeusser, Maria Kamenova,
  Konstantin Dimitriadis, Thomas Klopstock, M~Strupp, Thomas Brandt, and Klaus
  Jahn.
\newblock Locomotion speed determines gait variability in cerebellar ataxia and
  vestibular failure.
\newblock {\em Movement disorders}, 27(1):125--131, 2012.

\bibitem{wang2014stepping}
Yang Wang and Manoj Srinivasan.
\newblock Stepping in the direction of the fall: the next foot placement can be
  predicted from current upper body state in steady-state walking.
\newblock {\em Biology letters}, 10(9):20140405, 2014.

\bibitem{kurz2010levodopa}
Max~J Kurz and Jyhgong~Gabriel Hou.
\newblock Levodopa influences the regularity of the ankle joint kinematics in
  individuals with parkinson’s disease.
\newblock {\em Journal of computational neuroscience}, 28(1):131--136, 2010.

\bibitem{allert2001effects}
N~Allert, J~Volkmann, S~Dotse, H~Hefter, V~Sturm, and H-J Freund.
\newblock Effects of bilateral pallidal or subthalamic stimulation on gait in
  advanced parkinson's disease.
\newblock {\em Movement disorders: official journal of the Movement Disorder
  Society}, 16(6):1076--1085, 2001.

\bibitem{wen2019online}
Yue Wen, Jennie Si, Andrea Brandt, Xiang Gao, and He~Huang.
\newblock Online reinforcement learning control for the personalization of a
  robotic knee prosthesis.
\newblock {\em IEEE transactions on cybernetics}, 2019.

\bibitem{zhang2017human}
Juanjuan Zhang, Pieter Fiers, Kirby~A Witte, Rachel~W Jackson, Katherine~L
  Poggensee, Christopher~G Atkeson, and Steven~H Collins.
\newblock Human-in-the-loop optimization of exoskeleton assistance during
  walking.
\newblock {\em Science}, 356(6344):1280--1284, 2017.

\bibitem{alexander1997minimum}
R~McN Alexander.
\newblock A minimum energy cost hypothesis for human arm trajectories.
\newblock {\em Biological cybernetics}, 76(2):97--105, 1997.

\bibitem{liu2007evidence}
Dan Liu and Emanuel Todorov.
\newblock Evidence for the flexible sensorimotor strategies predicted by
  optimal feedback control.
\newblock {\em Journal of Neuroscience}, 27(35):9354--9368, 2007.

\bibitem{kording2007causal}
Konrad~P K{\"o}rding, Ulrik Beierholm, Wei~Ji Ma, Steven Quartz, Joshua~B
  Tenenbaum, and Ladan Shams.
\newblock Causal inference in multisensory perception.
\newblock {\em PLoS one}, 2(9):e943, 2007.

\bibitem{ackermann2010optimality}
Marko Ackermann and Antonie~J Van~den Bogert.
\newblock Optimality principles for model-based prediction of human gait.
\newblock {\em Journal of biomechanics}, 43(6):1055--1060, 2010.

\bibitem{seethapathi2015metabolic}
Nidhi Seethapathi and Manoj Srinivasan.
\newblock The metabolic cost of changing walking speeds is significant, implies
  lower optimal speeds for shorter distances, and increases daily energy
  estimates.
\newblock {\em Biology letters}, 11(9):20150486, 2015.

\bibitem{maxwell2001mechanical}
J~Maxwell~Donelan, Rodger Kram, and Kuo Arthur~D.
\newblock Mechanical and metabolic determinants of the preferred step width in
  human walking.
\newblock {\em Proceedings of the Royal Society of London. Series B: Biological
  Sciences}, 268(1480):1985--1992, 2001.

\bibitem{bertram2001multiple}
John~EA Bertram and Andy Ruina.
\newblock Multiple walking speed--frequency relations are predicted by
  constrained optimization.
\newblock {\em Journal of theoretical Biology}, 209(4):445--453, 2001.

\bibitem{bruijn2009slow}
Sjoerd~M Bruijn, Jaap~H van Die{\"e}n, Onno~G Meijer, and Peter~J Beek.
\newblock Is slow walking more stable?
\newblock {\em Journal of biomechanics}, 42(10):1506--1512, 2009.

\bibitem{dean2007effect}
Jesse~C Dean, Neil~B Alexander, and Arthur~D Kuo.
\newblock The effect of lateral stabilization on walking in young and old
  adults.
\newblock {\em IEEE Transactions on Biomedical Engineering}, 54(11):1919--1926,
  2007.

\bibitem{vejdani2013bio}
HR~Vejdani, Y~Blum, MA~Daley, and JW~Hurst.
\newblock Bio-inspired swing leg control for spring-mass robots running on
  ground with unexpected height disturbance.
\newblock {\em Bioinspiration \& biomimetics}, 8(4):046006, 2013.

\bibitem{chestnutt2007locomotion}
Joel Chestnutt, Philipp Michel, James Kuffner, and Takeo Kanade.
\newblock Locomotion among dynamic obstacles for the honda asimo.
\newblock In {\em 2007 IEEE/RSJ International Conference on Intelligent Robots
  and Systems}, pages 2572--2573. IEEE, 2007.

\bibitem{bhounsule2012design}
Pranav~A Bhounsule, Jason Cortell, and Andy Ruina.
\newblock Design and control of ranger: an energy-efficient, dynamic walking
  robot.
\newblock In {\em Adaptive Mobile Robotics}, pages 441--448. World Scientific,
  2012.

\bibitem{ding2014multi}
Ye~Ding, Ignacio Galiana, Alan Asbeck, Brendan Quinlivan, Stefano Marco~Maria
  De~Rossi, and Conor Walsh.
\newblock Multi-joint actuation platform for lower extremity soft exosuits.
\newblock In {\em 2014 IEEE International Conference on Robotics and Automation
  (ICRA)}, pages 1327--1334. Ieee, 2014.

\bibitem{johnson2007potential}
Michelle~J Johnson, Xin Feng, Laura~M Johnson, and Jack~M Winters.
\newblock Potential of a suite of robot/computer-assisted motivating systems
  for personalized, home-based, stroke rehabilitation.
\newblock {\em Journal of NeuroEngineering and Rehabilitation}, 4(1):6, 2007.

\bibitem{barliya2013expression}
Avi Barliya, Lars Omlor, Martin~A Giese, Alain Berthoz, and Tamar Flash.
\newblock Expression of emotion in the kinematics of locomotion.
\newblock {\em Experimental brain research}, 225(2):159--176, 2013.

\bibitem{hicheur2013perception}
Halim Hicheur, Hideki Kadone, Julie Grezes, and Alain Berthoz.
\newblock Perception of emotional gaits using avatar animation of real and
  artificially synthesized gaits.
\newblock In {\em 2013 Humaine Association Conference on Affective Computing
  and Intelligent Interaction}, pages 460--466. IEEE, 2013.

\bibitem{lourens2010communicating}
Tino Lourens, Roos Van~Berkel, and Emilia Barakova.
\newblock Communicating emotions and mental states to robots in a real time
  parallel framework using laban movement analysis.
\newblock {\em Robotics and Autonomous Systems}, 58(12):1256--1265, 2010.

\bibitem{more2010scaling}
Heather~L More, John~R Hutchinson, David~F Collins, Douglas~J Weber, Steven~KH
  Aung, and J~Maxwell Donelan.
\newblock Scaling of sensorimotor control in terrestrial mammals.
\newblock {\em Proceedings of the Royal Society B: Biological Sciences},
  277(1700):3563--3568, 2010.

\bibitem{usherwood2008compass}
James~R Usherwood, Katie~L Szymanek, and Monica~A Daley.
\newblock Compass gait mechanics account for top walking speeds in ducks and
  humans.
\newblock {\em Journal of Experimental Biology}, 211(23):3744--3749, 2008.

\bibitem{lin2011human}
Y~Chih Lin, B~Shiang Yang, Yu~Tzu Lin, Yi~Ting Yang, et~al.
\newblock Human recognition based on kinematics and kinetics of gait.
\newblock {\em Journal of Medical and Biological Engineering}, 31(4):255--263,
  2011.

\bibitem{neverova2016learning}
Natalia Neverova, Christian Wolf, Griffin Lacey, Lex Fridman, Deepak Chandra,
  Brandon Barbello, and Graham Taylor.
\newblock Learning human identity from motion patterns.
\newblock {\em IEEE Access}, 4:1810--1820, 2016.

\bibitem{wei2018behavioral}
Kunlin Wei and Konrad~Paul Kording.
\newblock Behavioral tracking gets real.
\newblock {\em Nature neuroscience}, 21(9):1146, 2018.

\bibitem{chambers2019pose}
Claire Chambers, Gaiqing Kong, Kunlin Wei, and Konrad Kording.
\newblock Pose estimates from online videos show that side-by-side walkers
  synchronize movement under naturalistic conditions.
\newblock {\em PloS one}, 14(6):e0217861, 2019.

\bibitem{ionescu2013human3}
Catalin Ionescu, Dragos Papava, Vlad Olaru, and Cristian Sminchisescu.
\newblock Human3. 6m: Large scale datasets and predictive methods for 3d human
  sensing in natural environments.
\newblock {\em IEEE transactions on pattern analysis and machine intelligence},
  36(7):1325--1339, 2013.

\bibitem{lin2014microsoft}
Tsung-Yi Lin, Michael Maire, Serge Belongie, James Hays, Pietro Perona, Deva
  Ramanan, Piotr Doll{\'a}r, and C~Lawrence Zitnick.
\newblock Microsoft coco: Common objects in context.
\newblock In {\em European conference on computer vision}, pages 740--755.
  Springer, 2014.

\bibitem{andriluka2018posetrack}
Mykhaylo Andriluka, Umar Iqbal, Eldar Insafutdinov, Leonid Pishchulin, Anton
  Milan, Juergen Gall, and Bernt Schiele.
\newblock Posetrack: A benchmark for human pose estimation and tracking.
\newblock In {\em Proceedings of the IEEE Conference on Computer Vision and
  Pattern Recognition}, pages 5167--5176, 2018.

\bibitem{hong2009kinematic}
Minna Hong, Joel~S Perlmutter, and Gammon~M Earhart.
\newblock A kinematic and electromyographic analysis of turning in people with
  parkinson disease.
\newblock {\em Neurorehabilitation and neural repair}, 23(2):166--176, 2009.

\bibitem{sun2019deep}
Ke~Sun, Bin Xiao, Dong Liu, and Jingdong Wang.
\newblock Deep high-resolution representation learning for human pose
  estimation.
\newblock {\em arXiv preprint arXiv:1902.09212}, 2019.

\bibitem{yang2017learning}
Wei Yang, Shuang Li, Wanli Ouyang, Hongsheng Li, and Xiaogang Wang.
\newblock Learning feature pyramids for human pose estimation.
\newblock In {\em Proceedings of the IEEE International Conference on Computer
  Vision}, pages 1281--1290, 2017.

\bibitem{newell2016stacked}
Alejandro Newell, Kaiyu Yang, and Jia Deng.
\newblock Stacked hourglass networks for human pose estimation.
\newblock In {\em European Conference on Computer Vision}, pages 483--499.
  Springer, 2016.

\bibitem{yang2018parsing}
Lu~Yang, Qing Song, Zhihui Wang, and Ming Jiang.
\newblock Parsing r-cnn for instance-level human analysis.
\newblock {\em arXiv preprint arXiv:1811.12596}, 2018.

\bibitem{kocabas2019self}
Muhammed Kocabas, Salih Karagoz, and Emre Akbas.
\newblock Self-supervised learning of 3d human pose using multi-view geometry.
\newblock {\em arXiv preprint arXiv:1903.02330}, 2019.

\bibitem{sun2018integral}
Xiao Sun, Chuankang Li, and Stephen Lin.
\newblock An integral pose regression system for the eccv2018 posetrack
  challenge.
\newblock {\em arXiv preprint arXiv:1809.06079}, 2018.

\bibitem{zhou2017unsupervised}
Tinghui Zhou, Matthew Brown, Noah Snavely, and David~G Lowe.
\newblock Unsupervised learning of depth and ego-motion from video.
\newblock In {\em Proceedings of the IEEE Conference on Computer Vision and
  Pattern Recognition}, pages 1851--1858, 2017.

\bibitem{vijayanarasimhan2017sfm}
Sudheendra Vijayanarasimhan, Susanna Ricco, Cordelia Schmid, Rahul Sukthankar,
  and Katerina Fragkiadaki.
\newblock Sfm-net: Learning of structure and motion from video.
\newblock {\em arXiv preprint arXiv:1704.07804}, 2017.

\bibitem{alp2018densepose}
R{\i}za Alp~G{\"u}ler, Natalia Neverova, and Iasonas Kokkinos.
\newblock Densepose: Dense human pose estimation in the wild.
\newblock In {\em Proceedings of the IEEE Conference on Computer Vision and
  Pattern Recognition}, pages 7297--7306, 2018.

\bibitem{goodman2013data}
Joseph~K Goodman, Cynthia~E Cryder, and Amar Cheema.
\newblock Data collection in a flat world: The strengths and weaknesses of
  mechanical turk samples.
\newblock {\em Journal of Behavioral Decision Making}, 26(3):213--224, 2013.

\bibitem{ionescu2014human3}
Catalin Ionescu, Dragos Papava, Vlad Olaru, and Cristian Sminchisescu.
\newblock Human3. 6m: Large scale datasets and predictive methods for 3d human
  sensing in natural environments.
\newblock {\em IEEE transactions on pattern analysis and machine intelligence},
  36(7):1325--1339, 2014.

\bibitem{Detectron2018}
Ross Girshick, Ilija Radosavovic, Georgia Gkioxari, Piotr Doll\'{a}r, and
  Kaiming He.
\newblock Detectron.
\newblock \url{https://github.com/facebookresearch/detectron}, 2018.

\bibitem{schache2009biomechanical}
Anthony~G Schache, Tim~V Wrigley, Richard Baker, and Marcus~G Pandy.
\newblock Biomechanical response to hamstring muscle strain injury.
\newblock {\em Gait \& posture}, 29(2):332--338, 2009.

\bibitem{chao1978graphical}
EY~Chao and KN~An.
\newblock Graphical interpretation of the solution to the redundant problem in
  biomechanics.
\newblock {\em Journal of Biomechanical Engineering}, 100(3):159--167, 1978.

\bibitem{stergiou2002frequency}
Nicholas Stergiou, Giannis Giakas, Jennifer~E Byrne, and Valerie Pomeroy.
\newblock Frequency domain characteristics of ground reaction forces during
  walking of young and elderly females.
\newblock {\em Clinical Biomechanics}, 17(8):615--617, 2002.

\bibitem{Weik2001}
Martin~H. Weik.
\newblock {\em Nyquist theorem}, pages 1127--1127.
\newblock Springer US, Boston, MA, 2001.

\bibitem{shinkai2008ball}
Hironari Shinkai, Hiroyuki Nunome, Yasuo Ikegami, and Masanori Isokawa.
\newblock Ball--foot interaction in impact phase of instep soccer kicking.
\newblock {\em Science and football VI}, 6:41, 2008.

\bibitem{mcmanus2010children}
Alison~M McManus, Eva~YW Chu, Clare~CW Yu, and Yong Hu.
\newblock How children move: activity pattern characteristics in lean and obese
  chinese children.
\newblock {\em Journal of obesity}, 2011, 2010.

\bibitem{ren2008whole}
Lei Ren, Richard~K Jones, and David Howard.
\newblock Whole body inverse dynamics over a complete gait cycle based only on
  measured kinematics.
\newblock {\em Journal of biomechanics}, 41(12):2750--2759, 2008.

\bibitem{dumas2007adjustments}
Raphael Dumas, Laurence Cheze, and J-P Verriest.
\newblock Adjustments to mcconville et al. and young et al. body segment
  inertial parameters.
\newblock {\em Journal of biomechanics}, 40(3):543--553, 2007.

\bibitem{hollman2007age}
John~H Hollman, Francine~M Kovash, Jared~J Kubik, and Rachel~A Linbo.
\newblock Age-related differences in spatiotemporal markers of gait stability
  during dual task walking.
\newblock {\em Gait \& posture}, 26(1):113--119, 2007.

\bibitem{akyol2007falls}
AD~Akyol.
\newblock Falls in the elderly: what can be done?
\newblock {\em International nursing review}, 54(2):191--196, 2007.

\bibitem{liederbach2014comparison}
Marijeanne Liederbach, Ian~J Kremenic, Karl~F Orishimo, Evangelos Pappas, and
  Marshall Hagins.
\newblock Comparison of landing biomechanics between male and female dancers
  and athletes, part 2: influence of fatigue and implications for anterior
  cruciate ligament injury.
\newblock {\em The American journal of sports medicine}, 42(5):1089--1095,
  2014.

\bibitem{osoba2017intelligence}
Osonde~A Osoba and William Welser~IV.
\newblock {\em An intelligence in our image: The risks of bias and errors in
  artificial intelligence}.
\newblock Rand Corporation, 2017.

\bibitem{robinovitch2013video}
Stephen~N Robinovitch, Fabio Feldman, Yijian Yang, Rebecca Schonnop, Pet~Ming
  Leung, Thiago Sarraf, Joanie Sims-Gould, and Marie Loughin.
\newblock Video capture of the circumstances of falls in elderly people
  residing in long-term care: an observational study.
\newblock {\em The Lancet}, 381(9860):47--54, 2013.

\bibitem{karayiannis2001extraction}
Nicolaos~B Karayiannis, Seshadri Srinivasan, Rishi Bhattacharya, Merrill~S
  Wise, James~D Frost, and Eli~M Mizrahi.
\newblock Extraction of motion strength and motor activity signals from video
  recordings of neonatal seizures.
\newblock {\em IEEE Transactions on medical imaging}, 20(9):965--980, 2001.

\bibitem{yi2014age}
Dong Yi, Zhen Lei, and Stan~Z Li.
\newblock Age estimation by multi-scale convolutional network.
\newblock In {\em Asian conference on computer vision}, pages 144--158.
  Springer, 2014.

\bibitem{xu2008hybrid}
Ziyi Xu, Li~Lu, and Pengfei Shi.
\newblock A hybrid approach to gender classification from face images.
\newblock In {\em 2008 19th International Conference on Pattern Recognition},
  pages 1--4. IEEE, 2008.

\bibitem{mendonca2018quantifying}
Rochelle Mendonca and Michelle Johnson.
\newblock Quantifying therapist--patient roles using video analysis during
  occupation-based therapy.
\newblock {\em American Journal of Occupational Therapy},
  72(4\_Supplement\_1):7211500013p1--7211500013p1, 2018.

\bibitem{cao2018openpose}
Zhe Cao, T~Simon, SE~Wei, and Y~Sheikh.
\newblock Openpose: real-time multi-person keypoint detection library for body,
  face, and hands estimation, 2018.

\bibitem{joukov2014online}
Vladimir Joukov, Michelle Karg, and Dana Kulic.
\newblock Online tracking of the lower body joint angles using imus for gait
  rehabilitation.
\newblock In {\em 2014 36th Annual International Conference of the IEEE
  Engineering in Medicine and Biology Society}, pages 2310--2313. IEEE, 2014.

\bibitem{delbracio2015burst}
Mauricio Delbracio and Guillermo Sapiro.
\newblock Burst deblurring: Removing camera shake through fourier burst
  accumulation.
\newblock In {\em Proceedings of the IEEE Conference on Computer Vision and
  Pattern Recognition}, pages 2385--2393, 2015.

\bibitem{mur2015orb}
Raul Mur-Artal, Jose Maria~Martinez Montiel, and Juan~D Tardos.
\newblock Orb-slam: a versatile and accurate monocular slam system.
\newblock {\em IEEE transactions on robotics}, 31(5):1147--1163, 2015.

\bibitem{mathis2018deeplabcut}
Alexander Mathis, Pranav Mamidanna, Kevin~M Cury, Taiga Abe, Venkatesh~N
  Murthy, Mackenzie~Weygandt Mathis, and Matthias Bethge.
\newblock Deeplabcut: markerless pose estimation of user-defined body parts
  with deep learning.
\newblock Technical report, Nature Publishing Group, 2018.

\bibitem{nath2019using}
Tanmay Nath, Alexander Mathis, An~Chi Chen, Amir Patel, Matthias Bethge, and
  Mackenzie~W Mathis.
\newblock Using deeplabcut for 3d markerless pose estimation across species and
  behaviors.
\newblock {\em Nature protocols}, 2019.

\end{thebibliography}

\end{document}